\def\year{2019}\relax
\documentclass[letterpaper]{article} %DO NOT CHANGE THIS
\usepackage{aaai19}  %Required
\usepackage{times}  %Required
\usepackage{helvet}  %Required
\usepackage{courier}  %Required
\usepackage{url}  %Required
\usepackage{graphicx}  %Required
\usepackage{color}

\usepackage{amsmath,amssymb} % define this before the line numbering.
\usepackage{booktabs}
\usepackage{bm}
\usepackage{array}

\usepackage{algorithm}
\usepackage{algorithmic}

\begin{document}
\title{Manifold-valued Image Generation with Wasserstein Generative Adversarial Nets}
\author{Zhiwu Huang$^\dagger$, Jiqing Wu$^\dagger$, Luc Van Gool$^{\dagger\ddagger}$\\
	$^\dagger$Computer Vision Lab, ETH Zurich, Switzerland \quad $^\ddagger$VISICS, KU Leuven, Belgium\\
	{\tt\small \{zhiwu.huang, jwu, vangool\}@vision.ee.ethz.ch}}
\maketitle
\begin{abstract}
Generative modeling over natural images is one of the most fundamental machine learning problems. However, few modern generative models, including Wasserstein Generative Adversarial Nets (WGANs), are studied on manifold-valued images that are frequently encountered in real-world applications. To fill the gap, this paper first formulates the problem of generating manifold-valued images and exploits three typical instances: hue-saturation-value (HSV) color image generation, chromaticity-brightness (CB) color image generation, and diffusion-tensor (DT) image generation. For the proposed generative modeling problem, we then introduce a theorem of optimal transport to derive a new Wasserstein distance of data distributions on complete manifolds, enabling us to achieve a tractable objective under the WGAN framework. In addition, we recommend three benchmark datasets that are CIFAR-10 HSV/CB color images, ImageNet HSV/CB color images, UCL DT image datasets. On the three datasets, we experimentally demonstrate the proposed manifold-aware WGAN model can generate more plausible manifold-valued images than its competitors.

\end{abstract}

\section{Introduction}

Building generative models of natural images has been a fundamental problem in machine learning. Such generative models are expected to estimate the probability distributions over the natural images. One of the most striking techniques is the family of generative adversarial networks (GANs) such as \cite{goodfellow2014generative,radford2015unsupervised,zhao2016energy,mao2016least}. The state-of-the-art GANs like \cite{arjovsky2017wasserstein,gulrajani2017improved,wei2018improving,miyato2018spectral} are good at approximating the distributions of Euclidean-valued (or real-valued) images implicitly by optimizing Wasserstein distance between the distributions of generated images and real images under an adversarial training framework.

While the GAN techniques have made great success for real-valued image generation, they are rarely applied to manifold-valued images that are of much interest in a variety of applications. For example, one of the most popular applications is processing phase-valued images, whose data live on either the Cycle $\mathbb{S}^1$ or the Sphere $\mathbb{S}^2$, in hue-saturation-value (HSV) and chromaticity-brightness (CB) color spaces. Since such HSV/CB spaces are more adapted to human color perception than the RGB space, many works like \cite{bergmann2014second,bergmann2015inpainting,bansal2015active,bergmann2016second,bacak2016second,laus2017nonlocal} have studied that the image processing models based on HSV/CB components can surpass the competitors developed in the RGB space. Analogously, for a better understanding of color semantics, producing HSV/CB images in an unsupervised manner would be a good alternative for the regular image generation over the RGB space. Another good application is for diffusion-tensor magnetic resonance imaging (DT-MRI). In many DT-MRI works like \cite{pennec2006riemannian,arsigny2007geometric,hasan2011review,jayasumana2013kernel,priya2015denoising}, DT images are generally processed on the Riemannian manifold of symmetric positive definite $3 \times 3$ matrices $\text{SPD}(3)$. In practice, researchers often have a severe lack of DT images for better analysis. Hence, generating photo-realistic DT images would have a high potential to benefit this field.

With the motivation in mind, we focus on generative modeling over manifold-valued images. In general, it proposes new challenges.
One of the most important issues is the generalization of the classic distribution distance to manifolds, while the other major challenge is the exploration on the manifold setting of the regular GAN objective. To address the first issue, we leverage the well-studied optimal transport (OT) theory \cite{fathi2010optimal,loeper2011regularity,de2014monge,fitschen2017optimal} on manifolds to introduce a new Wasserstein distance on complete manifolds. By adopting the proposed Wasserstein distance, we exploit a new model of Wasserstein GANs \cite{arjovsky2017wasserstein,gulrajani2017improved,wei2018improving} for manifold-valued image generation. Without loss of generality, we take three typical types of manifold-valued images for study, and suggest the proposed manifold-aware Wasserestein GAN approach to produce plausible samples residing on such manifolds. In summary, the paper gives rise to two original contributions:
	\begin{itemize}
		\item To the best of our knowledge, this paper proposes the novel problem of unsupervised manifold-valued image generation for the first time in the existing literature. 
		
		\item We generalize the Wasserestein GAN methodology to the Riemannian manifold setting so as to address the proposed problem of manifold-valued image generation.
	\end{itemize}

	%------------------------------------------------------------------------
	\section{Background}
	
	\begin{figure}[t]
		\begin{center}
			\includegraphics[width=0.8\linewidth]{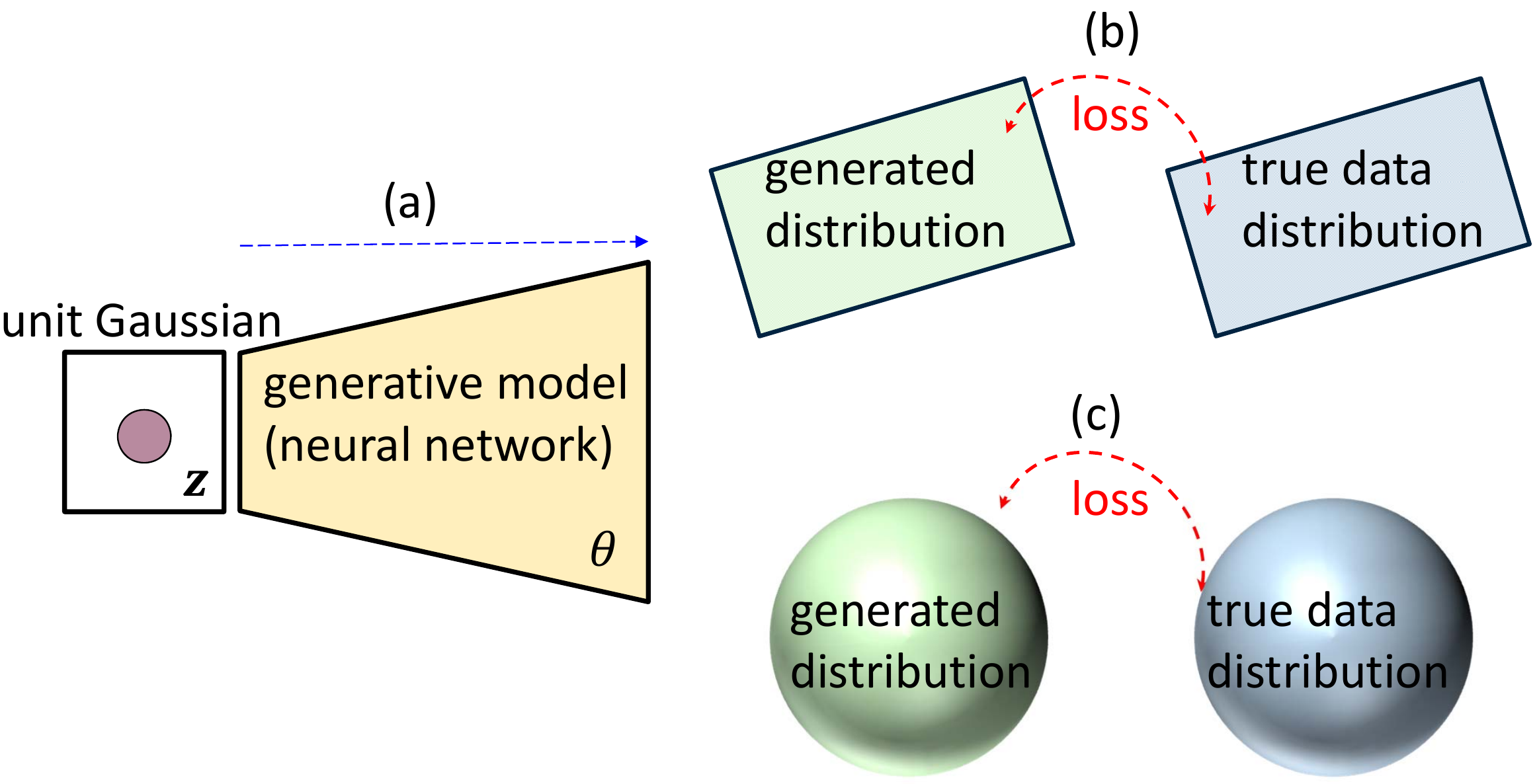}
		\end{center}
		\caption{Conceptual illustration of real-valued image generation (a)$\rightarrow$(b) where the data lie in Euclidean spaces, and manifold-valued image generation (a)$\rightarrow$(c) where the data reside on Riemannian manifolds.}
		\label{fig:long}
		\label{Fig1}
	\end{figure}
	
	\subsection{Real-valued Image Generation}
	Natural images are examples of what our visual world looks like and we refer to these as “samples from the true data distribution” $\mathbb{P}_r$. Typically, this kind of image values are Euclidean-valued (or real-valued). In other words, their geometrical structure can be respected well with using classical Euclidean metric. Accordingly, the image generation problem is concerned with learning the true data probability distribution. The classical solution is to learn a probability density, which may not exist in practice. In contrast, there also exists another typical approach that directly generates samples following a certain distribution $\mathbb{P}_g$, which approximates the true data distribution. As studied in \cite{arjovsky2017wasserstein} it is useful in two ways. First of all, this approach is able to represent distributions confined to a low dimensional manifold, which is easier to estimate. Second, the capability to generate samples easily is often more useful than estimating the numerical value of the density.
	
	Formally, one can consider a dataset of examples $\bm{x}_1,\ldots,\bm{x}_n$ sampled from a real data distribution $\mathbb{P}_r(\bm{x})$. As shown in Fig.\ref{Fig1} (a)$\rightarrow$(b), the blue region shows the part of the image space that, with a high probability consists of real images, and the elements in the space indicate the data points (each is one image). Now, the model also represents a distribution $\mathbb{P}_g (\bm{x})$ (green) that is defined implicitly by taking points from a unit Gaussian distribution (red) and mapping them through a neural network named the generative model (yellow). The network is a function with parameters $\bm{\theta}$, and learning these parameters will learn the generated distribution of images. The goal is then to seek parameters $\bm{\theta}$ that produce a distribution that matches the real data distribution closely. 
	Therefore, the core of the generation process is how to make the distribution of generated data close to the true data distribution, that is how to define the distribution distance. One of the most important measurements is Wasserstein-1 or Earth-Mover distance 
	\begin{equation}
	W(\mathbb{P}_r, \mathbb{P}_g) = \inf_{\gamma \in \Pi(\mathbb{P}_r, \mathbb{P}_g)} \mathbb{E}_{(\bm{x}, \bm{y})\sim \gamma}  [\|\bm{x} - \bm{y}\|],
	\label{Eq1}
	\end{equation}
	where $\Pi(\mathbb{P}_r, \mathbb{P}_g)$ denotes the set of all joint distributions $\gamma(\bm{x}, \bm{y})$ whose marginals are $\mathbb{P}_r, \mathbb{P}_g$ respectively. Intuitively, 
	$\gamma(\bm{x}, \bm{y})$ indicates how much mass should be transported from $\bm{x}$ to $\bm{y}$ in order to transform the distributions $\mathbb{P}_r$
	into the distribution $\mathbb{P}_g$. The Wasserstein distance then is the cost of the optimal transport plan.

	\subsection{Wasserstein GANs}
	For real-valued image generation, the state-of-the-art techniques are Generative Adversarial Networks (GANs) \cite{goodfellow2014generative,radford2015unsupervised,zhao2016energy,mao2016least}. The standard GAN framework establishes a min-max adversarial game between two competing networks. The generator
	network $G$ maps a source of noise to the input space. The discriminator network $D$ receives either a generated sample or a true data sample and must distinguish between the two. The generator is trained to fool the discriminator. 
	Theoretically, the original GAN framework actually minimizes Jensen-Shannon divergence between the true data distribution and generated sample distribution.
	
	By contrast, Wasserstein GANs \cite{arjovsky2017wasserstein,gulrajani2017improved,wei2018improving,miyato2018spectral} studied that minimizing a reasonable approximation of the Wasserstein-1 distance is able to reach the state-of-the-art GANs.
	To approximate the Wasserstein-1 distance, the original Wasserstein GAN imposed weight clipping constraints on the critic (referred to as the discriminator pre-Wasserstein) such that the optimal map of the discriminator is Lipschitz continuous. However, as proved in \cite{gulrajani2017improved}, the set of functions satisfying this constraint is merely a subset of the $k$-Lipschitz functions for some $k$ which depends on the clipping threshold and the critic architecture and thus inevitably causes some training failures. To address this issue, improved training of Wasserstein GAN \cite{gulrajani2017improved} enables a more stable GAN training by penalizing the norm of the gradient of the critic with respect to its inputs instead of clipping weights. In particular, this gradient penalty is simply added to the basic Wasserstein GAN loss for the following full objective:
	\begin{equation}
	\begin{aligned}
	\min_{G} \max_{D}  \mathbb{E}_{\bm{x} \sim \mathbb{P}_r} [D(\bm{x})] & -\mathbb{E}_{G(\bm{z}) \sim \mathbb{P}_g} [D(G(\bm{z}))]\\ & + \lambda \mathbb{E}_{\bm{\hat{x}} \sim \mathbb{P}_{\hat{x}}} [(\|\nabla_{\bm{\hat{x}}} D(\bm{\hat{x}})\|_2-1)^2],
	\end{aligned}
	\label{Eq2}
	\end{equation}
	where $\bm{z}$ is random noise, $\bm{\hat{x}}$ is random samples following the distribution $\mathbb{P}_{\hat{x}}$ that is sampled uniformly along straight lines between pairs of points sampled from $\mathbb{P}_r$ and $\mathbb{P}_g$, $\nabla_{\bm{\hat{x}}} D(\bm{\hat{x}})$ is the gradient with respect to $\bm{\hat{x}}$, $G(\cdot), D(\cdot)$ denotes the functions of generator and discriminator respectively.

	%------------------------------------------------------------------------
	
\section{Manifold-valued Image Generation}
	
	\subsection{Manifold-valued Images}

	In this paper, we concentrate on three typical instances of manifold-valued images, which are commonly encountered in computer vision and medical imaging.

	\textbf{HSV Images}: Each pixel in the HSV color model can be represented by a triple which specifies hue, saturation  and value respectively. As hue value is phase-based, the HSV data actually live on the product manifold of a Cyclic manifold and vector spaces $\mathcal{H} = \mathbb{S}^{1} \times [0,1]^2$.  Since it is known that both the Cyclic manifold $\mathbb{S}^{1}$ and the vector space $[0,1]^2$ are compact Riemannian manifolds, their product $\mathcal{H} = \mathbb{S}^{1} \times [0,1]^2$ is also a compact manifold.

	\textbf{CB Images}: Each pixel in the CB color model contains chromaticity and brightness components. Since our focus is on manifold-valued data, the whole paper studies the chromaticity (spherical) component lying on the compact manifold $\mathbb{S}^{2}$ for CB images.
	
	\textbf{DT Images}: In diffusion tensor (DT) images, each voxel is represented with a $3\times 3$ tensor, that is symmetric positive definite (SPD) matrix. Hence, the data of DT images reside on the manifold $\text{SPD}(3)$ of SPD matrices, which is known as a convex cone instead of a compact manifold.

	\subsection{Problem Formulation}
	
	In analogy to real-valued image generation, the task of manifold-valued image generation is to synthesize samples respecting a certain distribution for learning the distribution of real manifold-valued data. As shown in Fig.\ref{Fig1} (a)$\rightarrow$(c), the manifold-valued data lie on Riemannian manifolds rather than a Euclidean space. As the definition of manifold-valued data distribution is different of that of real-valued data distribution, it is infeasible to apply the traditional Wasserstein distance Eqn.\ref{Eq1} directly to measure the distance of such non-Euclidean data distribution, and the traditional Wasserstein GANs \cite{arjovsky2017wasserstein,gulrajani2017improved,wei2018improving,miyato2018spectral} are very likely to fail. Accordingly, for the new image generation task, we should consider two critical problems: 1) the definition on the distribution of manifold-valued data, and 2) the generalization of the distribution (Wasserstein) distance to manifolds.

    Fortunately, several Riemannian geometry studies \cite{pennec2006riemannian,arsigny2006log,arsigny2007geometric,huang2015log} have addressed the first issue well. In particular, they introduce various Riemannian metrics to define probability density functions on the underlying Riemannian manifolds.
	Formally, let $\bm{x, y}$ be two points of the manifold that we consider as a local reference and $\bm{v}=\overrightarrow{\bm{yx}}$ a vector of the tangent space $T_{\bm{y}}\mathcal{M}$ at the point $\bm{y}$. The smoothly varying family of inner products in each tangent space is known as the Riemannian metric. According to the theory of second-order differential equations, there has one and only one geodesic starting from that point with the tangent vector. This allows us to span the curved manifold in the flat tangent space along the geodesics (think of rolling a sphere along its tangent plane at a given point). The geodesics going through the reference point $\bm{y}$ are transformed into straight lines and the distance along these geodesics is preserved (at least in a neighborhood of $\bm{y}$). The function that maps to each tangent vector $\bm{v} \in T_{\bm{y}}\mathcal{M}$ the point $\bm{x}$ of the manifold that is achieved by the geodesic starting at $\bm{y}$ with this tangent vector is named the exponential map. This map is defined in the whole tangent space $T_{\bm{y}}\mathcal{M}$ but it is generally one-to-one only locally around 0 in the tangent space (i.e., around the reference point $\bm{y}$ in the manifold). For mapping the manifold data to the tangent space that respects Euclidean geometry, we denote
    $\bm{v} = \log_{\bm{y}} (\bm{x})$ as the inverse of the exponential map: this is the smallest vector such that $\bm{x} = \exp_{\bm{y}} \bm{v}$.

    For the second issue, we suggest to generalize the Wasserstein-1 distance to Riemannian manifolds. In theory, we  need to first study whether there exists an optimal map for the manifold setting. For optimal mass transport on real-valued data,  \cite{evans1997partial,villani2008optimal,lei2017geometric} proved the existence of the optimal map, which has a convex potential (i.e. $D(\bm{x}) = \nabla \phi(\bm{x})$ with $\phi(\bm{x})$ being convex) and is shown to be the only map with the convex potential. Analogously, \cite{mccann2001polar,de2014monge,fathi2010optimal} also studied that there exists an optimal map for the manifold case. Furthermore, \cite{fathi2010optimal} proved that the optimal transport problem can be solved for the square Riemannian metric on any complete manifolds without any assumption on the compactness or curvature, with the usual restriction on the measures. The following theorem is given for the existence of the optimal map on complete manifolds. For its proof, we refer readers to \cite{fathi2010optimal} (mainly in Page 18-20).

    \vspace{3mm}
	
	\noindent \textbf{Theorem 1.} \emph{Let $\mathcal{M}$ be a Riemannian manifold, and consider the cost $c = d^2/2$, with $d$ being the Riemannian distance metric that is required to be complete. Given two probability distributions $\mathbb{P}_r$ and $\mathbb{P}_g$ supported on the manifold $\mathcal{M}$, there exists a convex potential function $\phi : \mathcal{M} \rightarrow \mathbb{R} \cup \{+\infty\}$ such that $D(\bm{x}) = \exp_{\bm{y}}(\nabla_g \phi(\bm{x}))$ is the unique optimal transport map sending $\mathbb{P}_r$ to $\mathbb{P}_g$, where $\nabla_g$ indicates the gradient with respect to the Riemannian metric $g=\langle \log_{\bm{y}}(\bm{x}_1), \log_{\bm{y}}(\bm{x}_2) \rangle_{\bm{y}}$ on the manifold $\mathcal{M}$.}

	\vspace{3mm}
	
    Following \textbf{Theorem 1}, we turn to study the complete Riemannain metrics for the existence of optimal transport map on our studied manifolds where the HSV, CB and DT image data reside on. It is known that any Riemannian metrics defined on compact Riemannian manifolds are complete. Therefore, for HSV and CB images, Riemannian metrics defined on them are all complete. As to the case of DT images, whose data reside on a non-compact manifold, we restore to employing the well-studied Log-Euclidean metric \cite{arsigny2006log,arsigny2007geometric,huang2015log}, which is actually an inner product distance and thus the resulting metric space is complete  \cite{minh2016covariance}. For a more general guidance to satisfy the completeness requirement, readers are suggested to follow the Hopf–Rinow theorem \cite{hopf1931begriff}.
	
	Consequently, inspired by the original Wasserstein distance Eqn.\ref{Eq1} in a Euclidean setting, we present a new Wasserstein distance on the underlying Riemannian manifold. By leveraging the complete Riemannian metric that generally applies classic logarithm map to yield Euclidean representations, the new Wasserstein distance can be derived by:
	\begin{equation}
	W(\mathbb{P}_r, \mathbb{P}_g) = \inf_{\gamma \in \Pi(\mathbb{P}_r, \mathbb{P}_g)} \mathbb{E}_{(\bm{x}, \bm{y})\sim \gamma}  [\|\log_{\bm{y}}(\bm{x}) - \log_{\bm{y}}(\bm{y})\|],
	\label{Eq11}
	\end{equation}
	where $\bm{x}, \bm{y}$ are points on the Riemannian manifold, $\Pi(\mathbb{P}_r, \mathbb{P}_g)$ indicates the set of all joint distributions $\gamma(\bm{x}, \bm{y})$, and
	$\gamma(\bm{x}, \bm{y})$ denotes how much mass will be transported from $\bm{x}$ to $\bm{y}$.

	\vspace{3mm}
	
	\begin{figure}[t]
		\begin{center}
			\includegraphics[width=0.8\linewidth]{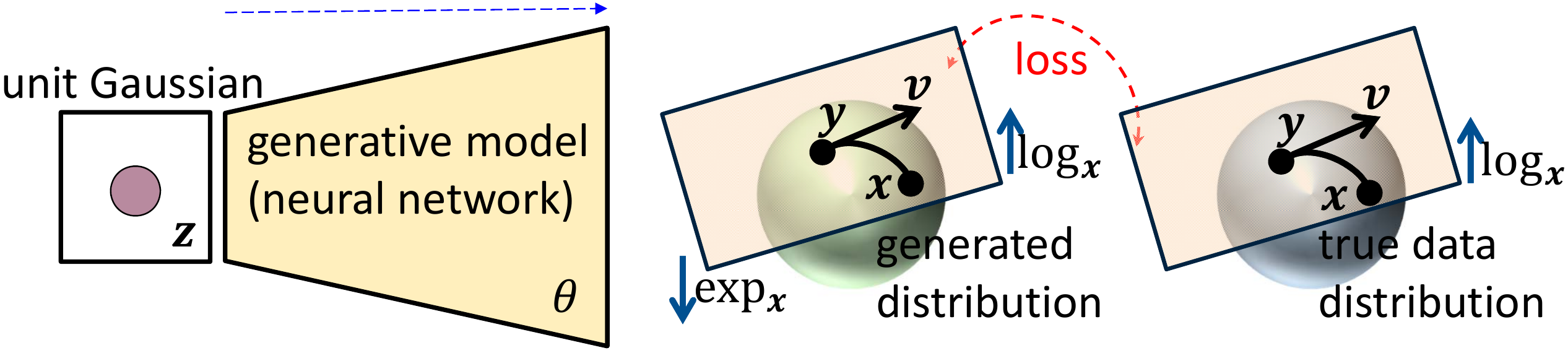}
		\end{center}
		\caption{Overview of the proposed manifold-aware Wasserstein GAN (manifoldWGAN) approach for manifold-valued image generation. The manifoldWGAN introduces logarithm map and exponential map between Riemannian manifold and tangent space to the setting of Wassersten GANs.}
		\label{fig:long}
		\label{Fig2}
	\end{figure}
	
	%------------------------------------------------------------------------
	\section{Manifold-aware Wasserstein GAN}
	
	According to \textbf{Theorem 1} and the derived Wasserstein distance Eqn.\ref{Eq11}, we generalize the objective function Eqn.\ref{Eq2} of Wasserstein GAN\footnote{As studying the generalization of all the Wasserstein GANs \cite{arjovsky2017wasserstein,gulrajani2017improved,wei2018improving} is beyond the scope of this paper, we choose the most representative Wasserstein GAN \cite{gulrajani2017improved} for study.} to Riemannian manifolds
	\begin{equation}
	\begin{aligned}
	\min_{G} \max_{D}  & \, \mathbb{E}_{\bm{x} \sim \mathbb{P}_r} [D(\log_{\bm{y}}(\bm{x}))] \\ & -\mathbb{E}_{G(\bm{z}) \sim \mathbb{P}_g} [D(\log_{\bm{y}}(\exp_{\bm{y}}(G(\bm{z}))))]\\ & + \lambda \mathbb{E}_{\bm{\hat{x}} \sim \mathbb{P}_{\bm{\hat{x}}}} [(\|\nabla_{\bm{\hat{x}}} D(\bm{\hat{x}})\|_2-1)^2],
	\end{aligned}
	\label{Eq3}
	\end{equation}
	where $\log_{\bm{y}}(\cdot)$ and $\exp_{\bm{y}}(\cdot)$ are the logarithm and exponential maps respectively for the underlying manifold. As the logarithm map $\log_{\bm{y}}(\cdot)$ projects the manifold-valued data to Euclidean space, any regular networks can be applied directly to the resulting data. To generate valid manifold-valued data using the generative network $G$, we employ exponential map $\exp_{\bm{y}}(\cdot)$ to transform the data back to the manifold. An overview of the proposed manifold-aware Wasserstein GAN (manifoldWGAN) approach is shown in Fig.\ref{Fig2}. As done in \cite{gulrajani2017improved}, $\mathbb{P}_{\bm{\hat{x}}}$ is also defined sampling uniformly along straight lines between pairs of points sampled from the data distribution $\mathbb{P}_r$ and the generator distribution $\mathbb{P}_g$. In the context of manifold-valued data, we apply logarithm and exponential maps to the linear sampling:
	\begin{equation}
	\begin{aligned}
	\bm{\hat{x}}= (1-t) \log_{\bm{y}}(\bm{x}) + t\log_{\bm{y}}(\exp_{\bm{y}}(G(\bm{z}))),
	\end{aligned}
	\label{Eq8}
	\end{equation}
	where $0 \leq t \leq 1$.
	
	In this paper, our focus is on the generation of HSV, CB and DT images. Therefore, we consider the data on the product manifold $\mathcal{H} = \mathbb{S}^{1} \times [0,1]^2$, the spherical data on $\mathbb{S}^{2}$ and the SPD data on $\text{SPD}(3)$. Specially, we study the typical forms for their resulting exponential map and logarithm map to achieve the objective Eqn.\ref{Eq3}.
	
	\vspace{3mm}

	\noindent \textbf{HSV case}:  \emph{Following \cite{bergmann2014second,bergmann2016second}, we adopt the representation system $\mathbb{S}^{1}  \cong [-\pi, \pi)$ to interpret any point $\bm{x} \in \mathcal{H} = \mathbb{S}^{1} \times [0,1]^2$ as consisting of two components $(\bm{x}_{\mathbb{S}}, \bm{x}_{\mathbb{R}})$. As a result, the corresponding logarithm and exponential maps can be derived as}
	\begin{equation}
	\begin{aligned}
	\log_{\bm{y}}(\bm{x}) & = [\text{mod}(\bm{x}_{\mathbb{S}}-\bm{y}_{\mathbb{S}}, 2\pi), \bm{x}_{\mathbb{R}}-\bm{y}_{\mathbb{R}}],\\
	\exp_{\bm{y}}(\bm{v}) & = [\text{mod}(\bm{v}_{\mathbb{S}}+\bm{y}_{\mathbb{S}}, 2\pi), \bm{v}_{\mathbb{R}}+\bm{y}_{\mathbb{R}}],
	\label{Eq10}
	\end{aligned}
	\end{equation}
	\emph{where $\text{mod}$ denotes the modulo operation.}
	
	\vspace{3mm}

	\noindent \textbf{CB case}:  \emph{As studied in \cite{bhattacharya2002nonparametic,lee2007dimensionality,bacak2016second,laus2017nonlocal}, on the sphere manifold $\mathbb{S}^n$, the logarithm and exponential maps can be expressed by}
	\begin{equation}
	\begin{aligned}
	\log_{\bm{y}}(\bm{x}) & = \frac{d(\bm{x},\bm{y})}{\|p_{\bm{y}}(\bm{x}-\bm{y})\|_F}p_{\bm{y}}(\bm{x}-\bm{y}),\\
	\exp_{\bm{y}}(\bm{v}) & = \cos(\|\bm{v}\|_F)\bm{y}+\frac{\sin(\|\bm{v}\|_F)}{\|\bm{v}\|_F}\bm{v},
	\label{Eq5}
	\end{aligned}
	\end{equation}
	\emph{where $\|\cdot\|_F$ indicates the  Frobenius norm operation, $p_{\bm{y}}(\bm{H})=\bm{H}-\text{trace}(\bm{y}^T\bm{H})\bm{y}$.}

	\vspace{3mm}
	
	\noindent \textbf{DT case}: \emph{\cite{arsigny2006log,arsigny2007geometric,huang2015log,huang2017riemannian} studied that the Log-Euclidean metric for the SPD manifold $\text{SPD}(n)$ is derived by employing the Lie group structure. The resulting logarithmic and exponential maps is expressed with matrix logarithms and exponential operations:}
	\begin{equation}
	\begin{aligned}
	\log_{\bm{y}}(\bm{x}) & = D_{\log(\bm{y})}\exp.(\log(\bm{y})-\log(\bm{x})),\\
	\exp_{\bm{y}}(\bm{v}) & = \exp(\log(\bm{y})+D_{\bm{y}}\log.\bm{v}),
	\label{Eq7}
	\end{aligned}
	\end{equation}
	\emph{where $D_{\log(\bm{y})}\exp.=(D_{\bm{y}}\log.)^{-1}$ is achieved by the differentiation of the equality $\log\circ\exp=\bm{I}$, and here $\bm{I}$ is the identity matrix.}

	\begin{algorithm}[t]
		\caption{Manifold-aware Wasserstein GAN (manifoldWGAN), our proposed algorithm. All the experiments in the paper used the default values $\lambda=10$,  $n_{\text{critic}}=5$.}
		\textbf{Require}: $\alpha$, learning rate. $m$, the batch size. $n_{\text{critic}}$, the critic iterations per generation iteration. $\lambda$, the balance parameter of gradient norm penalty, $\bm{w}_0$, initial critic parameters. $\bm{\theta}_0$, initial generator's parameter. \\
		1: \textbf{while} $\bm{\theta$} has not converged \textbf{do}\\
		2: \quad{}\textbf{for} t=0, \ldots, $n_{\text{critic}}$ \textbf{do}\\
		3: \quad{}\quad{}Sample $\{\bm{x}^{(i)}\}_{i=1}^{m} \sim \mathbb{P}_r$ a batch from the real data.\\
		%4. \quad{}\quad{}Apply logarithm map to transform the real data \\ 
		%\phantom{5.} \quad{}\quad{}$\{\bm{x}^{(i)}\}_{i=1}^{m}$ to Euclidean data $\{\log_{\bm{y}}{\bm{x}}^{(i)}\}_{i=1}^{m}$\\
		4: \quad{}\quad{}Sample $\{\bm{z}^{(i)}\}_{i=1}^{m} \sim \mathbb{P}_g$ a batch of prior samples.\\
		%5:\quad{}\quad{} $g_w \leftarrow \nabla_w[\frac{1}{m}\Sigma_{i=1}^m f_w(\bm{x}^{(i)})-\frac{1}{m}\Sigma_{i=1}^m f_w(g_{\theta}(\bm{z}^{(i)}))]$.\\
		5:\quad{}\quad{} $D_w \leftarrow\nabla_w[\mathcal{L}]$ where $\mathcal{L}$ is computed by Eqn.\ref{Eq3}.\\
		6:\quad{}\quad{} $\bm{w} \leftarrow \bm{w} + \alpha \cdot \text{AdamOptimizer}(\bm{w},D_w)$ \\
		7: \quad{}\textbf{end for}\\
		8: \quad{}Sample $\{\bm{z}^{(i)}\}_{i=1}^{m} \sim \mathbb{P}_g$ a batch of prior samples.\\
		9: \quad{}$G_{\theta} \leftarrow\nabla_{\theta}[-\frac{1}{m}\Sigma_{i=1}^m D_w(\log_{\bm{y}}(\exp_{\bm{y}}(G_{\theta}(\bm{z}^{(i)}))))]$\\
		10:\quad{}$\bm{\theta} \leftarrow \bm{\theta} + \alpha \cdot  \text{AdamOptimizer}(\bm{\theta},G_{\theta})$ \\
		11: \textbf{end while}
		\label{Alg1}
	\end{algorithm}
	
	\vspace{3mm}

Following such basic logarithm and exponential maps\footnote{For the HSV and CB case, the computational cost of $\log$ and $\exp$ operations are cheap as they only include simple operations like modulo, cosine and sine. For the DT case, the dominant computational complexity of the used $\log$ and $\exp$ operations lies on computing the matrix logarithm and exponential on SPD matrix valued DT image pixels. As the size of each SPD matrix is only of $3\times3$, the $\log$ and $\exp$ computation is also cheap for network training.} on the corresponding manifolds of the processed manifold-valued samples, the algorithm of the proposed manifoldWGAN is presented in Algorithm.\ref{Alg1}.

\section{Experiment}

For the studied manifold-valued image generation problem, we suggest three benchmark evaluations that use the HSV and CB images of the well-known CIFAR-10 \cite{krizhevsky2009learning}, ImageNet \cite{oord2016pixel}, and the popular UCL DT image dataset \cite{cook2006camino}. As our focus is on
evaluating the generative models for image generation, we mainly compare the proposed manifoldWGAN against the state-of-the-art GAN techniques including deep convolutional GAN (DCGAN) \cite{radford2015unsupervised}, least square GAN (LSGAN) \cite{mao2016least} and Wasserstein GAN (WGAN)\footnote{The official code is available at
\url{https://github.com/igul222/improved_wgan_training}} \cite{gulrajani2017improved} that has proved to improve the original WGAN \cite{arjovsky2017wasserstein}.

%\zhiwu{Network structure? ResNet?}
In all the evaluations, we follow \cite{gulrajani2017improved} to use residual network for all compared models. The setting of the employed residual networks is the same as the ones used in \cite{gulrajani2017improved}. Specially, we utilize pre-activation residual blocks with two 3$\times 3$ convolutional layers and ReLU nonlinearity. Some residual blocks perform downsampling (in the critic) using mean pooling after the second convolutional layer, or nearest-neighbor upsampling (in the generator) before the second convolution. As done by \cite{gulrajani2017improved}, we employ batch normalization in the generator but not the critic for WGAN and our proposed manifoldWGAN. We finally optimize the network using Adam with learning rate 0.0002, decayed linearly to 0 over 100K generator iterations, and batch size 64. For further architectural details, please refer to the official implementation of \cite{gulrajani2017improved}.

	\begin{figure*}[t!]
    \centering
    \begin{tabular}{c|c|c|c|c}
    \toprule
    & CIFAR-HSV & CIFAR-RGB & ImageNet-HSV & ImageNet-RGB \\ \hline
    
     WGAN & \includegraphics[width=0.15\linewidth]{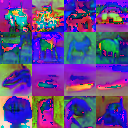}&
    \includegraphics[width=0.15\linewidth]{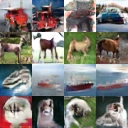}&
    \includegraphics[width=0.15\linewidth]{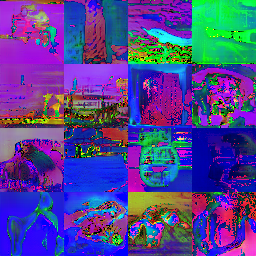}&
    \includegraphics[width=0.15\linewidth]{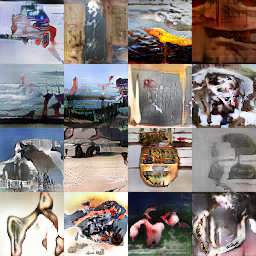}\\
      \hline
    
     \textbf{Proposed} &    \includegraphics[width=0.15\linewidth]{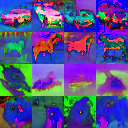}&
    \includegraphics[width=0.15\linewidth]{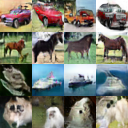}&
    \includegraphics[width=0.15\linewidth]{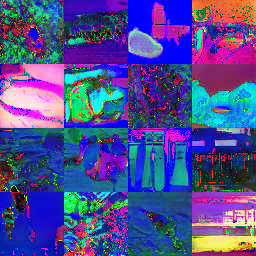}&
    \includegraphics[width=0.15\linewidth]{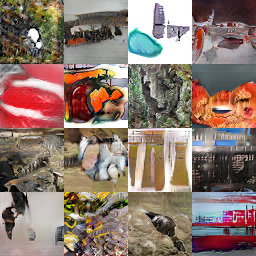}\\

    \midrule
    & CIFAR-CB & CIFAR-RGB & ImageNet-CB & ImageNet-RGB \\ \hline
    
     WGAN & \includegraphics[width=0.15\linewidth]{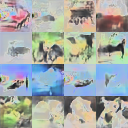}&
    \includegraphics[width=0.15\linewidth]{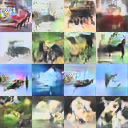}&
    \includegraphics[width=0.15\linewidth]{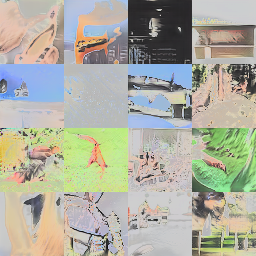}&
    \includegraphics[width=0.15\linewidth]{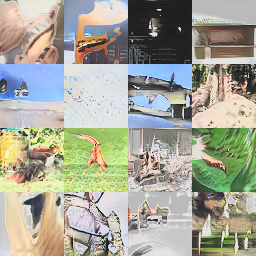}\\
      \hline

     \textbf{Proposed} & \includegraphics[width=0.15\linewidth]{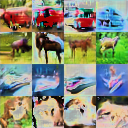}&
    \includegraphics[width=0.15\linewidth]{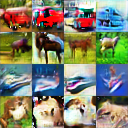}&
    \includegraphics[width=0.15\linewidth]{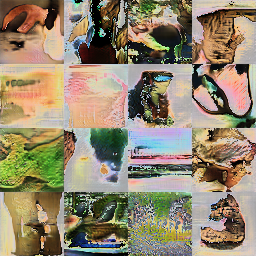}&
    \includegraphics[width=0.15\linewidth]{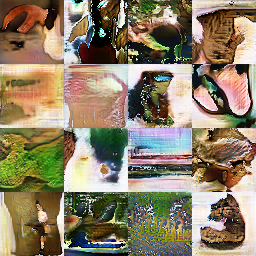} \\

    \bottomrule
    
    \end{tabular}
    \vspace{-0.2cm}
    \caption{Generated samples of WGAN \cite{gulrajani2017improved} and the proposed manifoldWGAN for HSV/CB image generation on CIFAR-10/ImageNet with translated RGB images. Please refer to supplementary material for more visual results.}
    \label{fig:samples-hsv-cb}
    \vspace{-0.2cm}
    \end{figure*}

\subsection{HSV Image Generation}
For HSV color image generation, we choose the CIFAR-10 and ImageNet datasets, both of which are standard benchmarks in real-valued image generation. The CIFAR-10 dataset consists of 60000 $32 \times 32$ colour images in 10 classes, with 6000 images per class. We use the $64 \times 64$ version of ImageNet, which contains 1,281,149 training images and 49,999 images for testing. To gather HSV color images whose values are on $\mathcal{H} = \mathbb{S}^{1} \times [0,1]^2$, we transfer the images from the RGB space to the HSV space. 

For generating valid elements on $\mathcal{H} = \mathbb{S}^{1} \times [0,1]^2$, the proposed manifoldWGAN suggests to use the corresponding logarithm and exponential maps Eqn.\ref{Eq10} (where each pixel value of the reference point $\bm{y}$ is set to $[\pi, 0,0]$\footnote{We empirically find varying the value of anchor point impacts the performance of our model slightly.}) during the optimization of the objective Eqn.\ref{Eq3}, while the state-of-the-art generation technique WGAN treats the data as real-valued data with its original Wasserstein-based GAN loss.

Fig.\ref{fig:samples-hsv-cb} qualitatively shows the generation results of WGAN and the proposed manifoldWGAN on CIFAR-10 and ImageNet. The results justify that our manifoldWGAN can generate more visually pleasing HSV images, whose resulting RGB images appear to be better in both terms of image quality and semantic. 

In addition, we follow \cite{heusel2017gans,lucic2017gans} to adopt Fr\'echet inception distance (FID)\footnote{The FID uses the statistics (i.e., mean and covariance) of real world samples and compare it to the statistics of synthetic samples.} to compare the WGAN and our manifoldWGAN quantitatively on the CIFAR-10 dataset. 
As the original FID metric is computed on RGB images\footnote{While it might be possible to adapt the FID computation on manifold-valued images, we leave it as one of future works due to the absence of effective manifold-valued inception model.}, we transfer the produced HSV images of generative models to the corresponding RGB images. Besides, we also present the FIDs of the state-of-the-art GANs \cite{radford2015unsupervised,mao2016least,gulrajani2017improved}\footnote{We also evaluated spectral normalized WGAN (SN-WGAN) \cite{miyato2018spectral} and our manifold-enhanced SN-WGAN. Due to the space limit, we only report that their FIDs are 31.3 and 24.2 resp. for the CIFAR-10 HSV case, showing our method can enable other WGANs to work well for manifold-valued image generation.} that are all originally designed for RGB images. The results reported in Tab.\ref{lab:fid} demonstrate our manifoldWGAN performs better than WGAN for HSV image generation on both of the CIFAR-10 and ImageNet datasets. More interestingly, the performances of our manifoldWGAN-HSV on such two datasets are also consistently better than those of the state-of-the-art GANs (e.g., WGAN) that work for RGB images directly. This is because our manifoldWGAN for HSV-based image generation targets a better understanding on color semantics. From this perspective, studying HSV-based image generation would be a good direction to enhance the regular image generation.

	\begin{figure}[t]
		\begin{center}
			\includegraphics[width=0.48\linewidth]{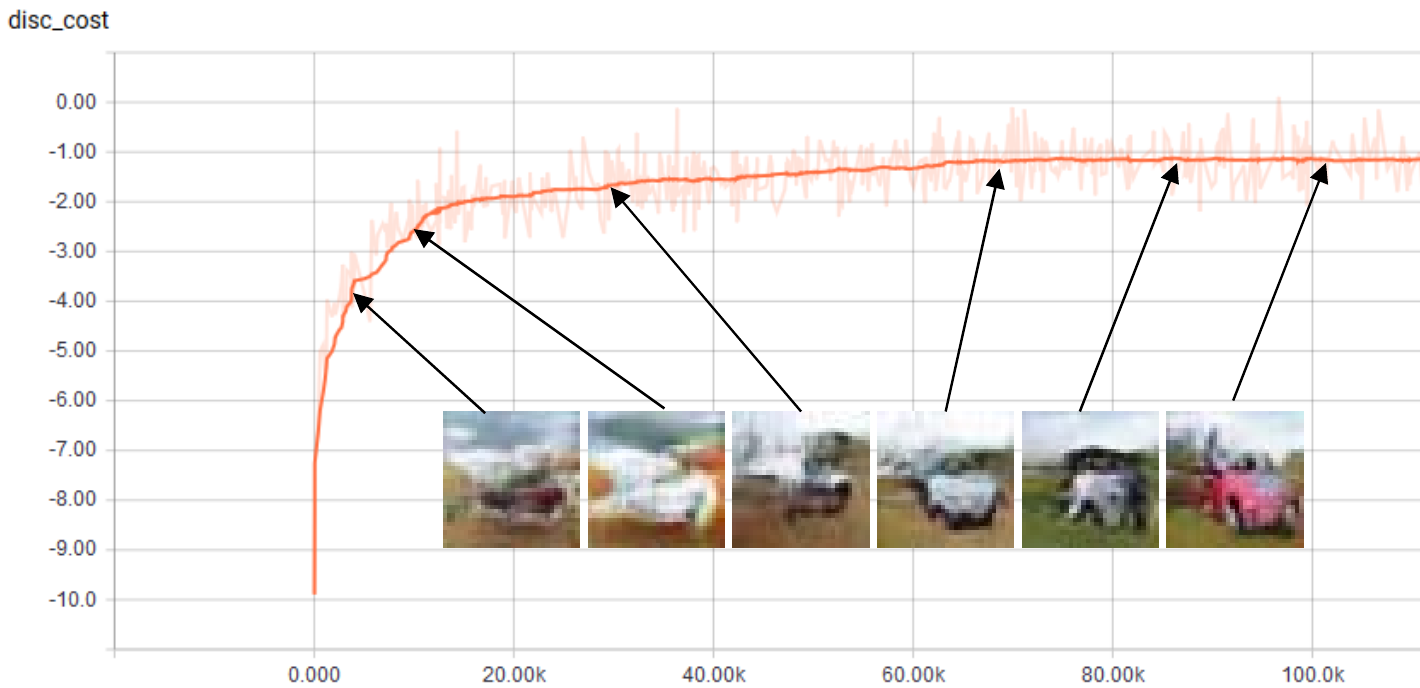}
			\includegraphics[width=0.48\linewidth]{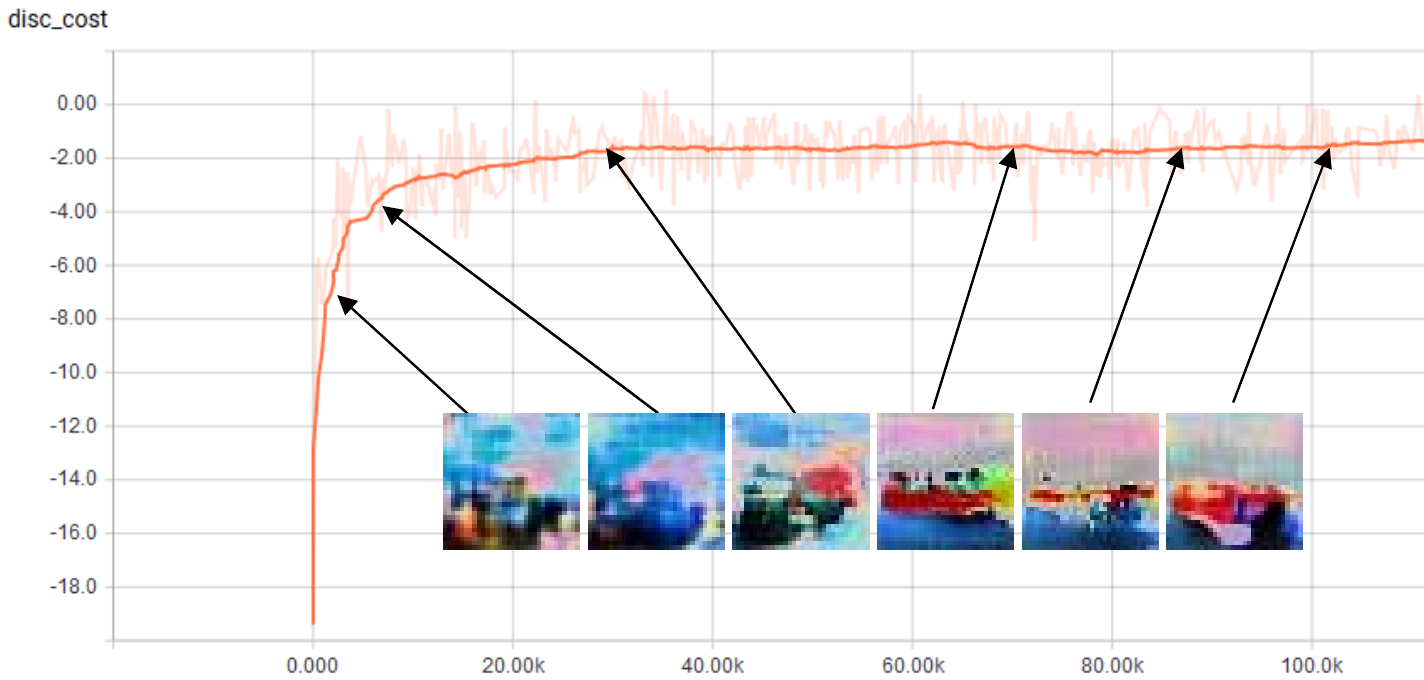}

		\end{center}
		\caption{Training curves and generated samples at different  training stages of our manfioldWGAN on CIFAR-10 HSV image set (left) and CB image set (right). The value on y-axis denotes the negative critic cost (Wasserstein distance), and the value on the x-axis indicates the iteration number.}
		\label{fig:conv_hsv_cb} 
	\end{figure}

We finally show the training curves of negative Wasserstein GAN loss (Wasserstein distance) and generated samples at different stages of training in Fig.\ref{fig:conv_hsv_cb}. As we can see, the loss decreases constistently as training progresses and sample quality increases. This verifies a strong correlation between lower loss and better sample quality.

	\begin{table}[t]
		%\linespread{1.3}
		\begin{center}
			\begin{tabular}{lcc}
				\toprule
				Method & CIFAR-10  & ImageNet \\
				\midrule
				% DCGAN-RGB \cite{radford2015unsupervised} & 37.7 & 89.5 \\
				DCGAN-RGB & 37.7 & 95.5 \\
				LSGAN-RGB & 31.9 & 82.4 \\
				WGAN-RGB & 29.3 & 71.2 \\
				\midrule
				WGAN-HSV & 38.4 & 76.7\\
				\textbf{manifoldWGAN-HSV} & \textbf{27.2} & \textbf{68.4}\\
				\midrule
				WGAN-CB & 80.7 & 101.3\\
				\textbf{manifoldWGAN-CB} & 59.4 & 89.2\\
				\bottomrule
			\end{tabular}
		\end{center}
		\caption{FIDs of compared methods performing generation on RGB images or HSV images of CIFAR-10 and ImageNet. The FIDs of all the HSV/CB-based models are computed by translating their produced HSV/CB images to RGB images.}
		\label{lab:fid}
	\end{table}

\subsection{CB Image Generation}
To evaluate our manifoldWGAN for image generation on the sphere manifold $\mathbb{S}^{2}$, we collect CB images from CIFAR-10 and ImageNet. The basic settings on them are the same with those of the last evaluation. For the task, we extract the chromaticity component of the CB images so that the generation problem on pure spherical data can be better studied.

As done in the last evaluation, we compare our manifoldWGAN against the state-of-the-art GAN technique WGAN. Our manifoldWGAN proposes to make use of the manifold geometry-aware logarithm and exponential maps Eqn.\ref{Eq5} (where each pixel value of the anchor point $\bm{y}$ is set to $[\frac{1}{\sqrt{3}},\frac{1}{\sqrt{3}},\frac{1}{\sqrt{3}}]$\footnote{The performance of our model changes slightly when varying the reference point.}) on the sphere manifold during the sample generation. As shown in Fig.\ref{fig:samples-hsv-cb}, we discover that the WGAN approach generates samples with low-chromaticity, while the proposed manifoldWGAN can generate higher-quality samples. This shows the clear advantage of our method.

\begin{figure*}[t]
		\begin{center}
		\begin{tabular}{c|c|c}
        \toprule
       Groundtruth & WGAN & \textbf{Proposed} \\ \hline
			\includegraphics[width=0.25\linewidth]{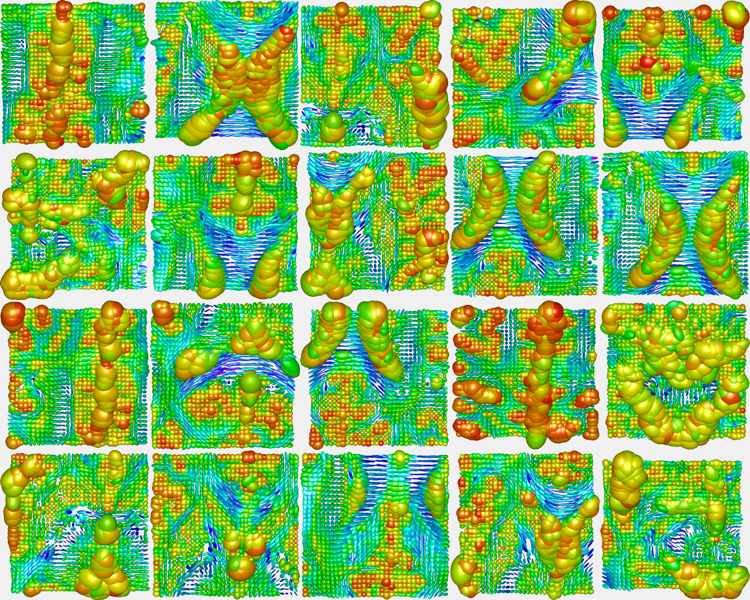} &
			\includegraphics[width=0.25\linewidth]{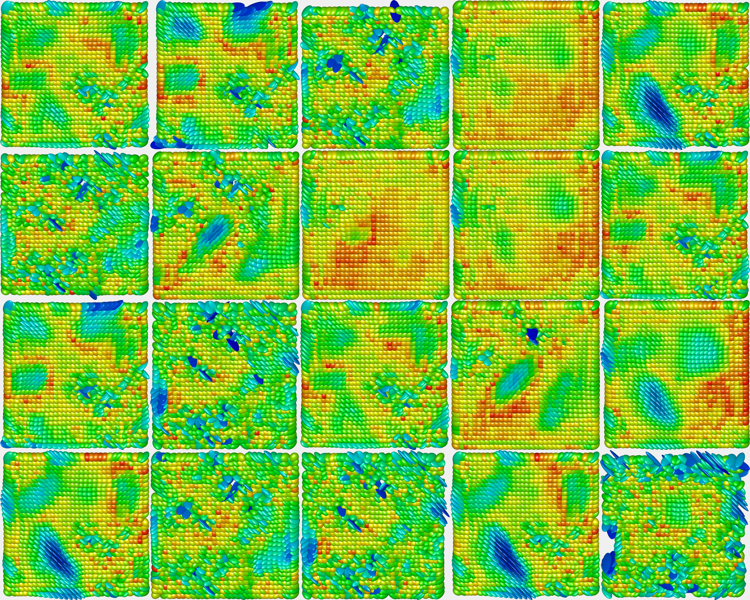} &
			\includegraphics[width=0.25\linewidth]{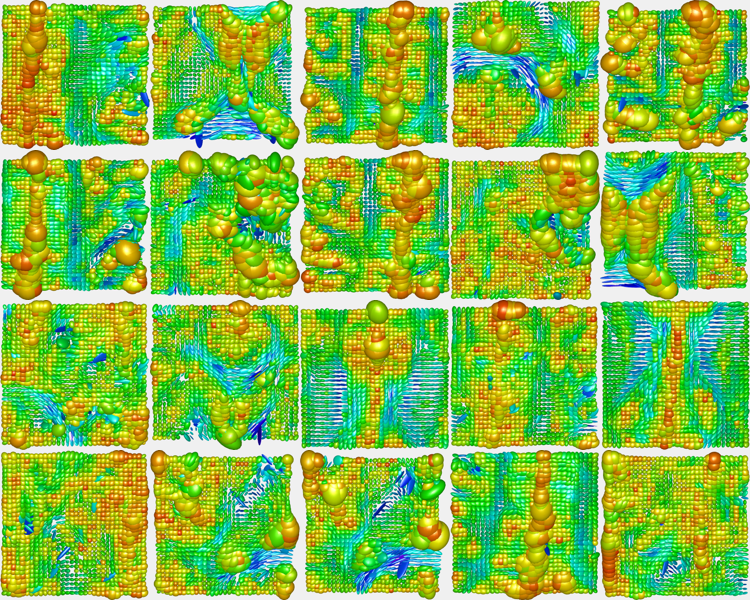}\\
			\bottomrule

		\end{tabular}
		\end{center}

		\caption{Groundtruth DT images (left) , Generated samples of WGAN \cite{gulrajani2017improved} (middle) and the proposed manifoldWGAN (right) for DT image generation on the UCL DT image dataset. The figure is better viewed via zooming in. For more visual results, please refer to supplementary material.}
		\label{fig:ucl-dt} 
	\end{figure*}

    \begin{figure}[t]
		\begin{center}
			\includegraphics[width=0.7\linewidth]{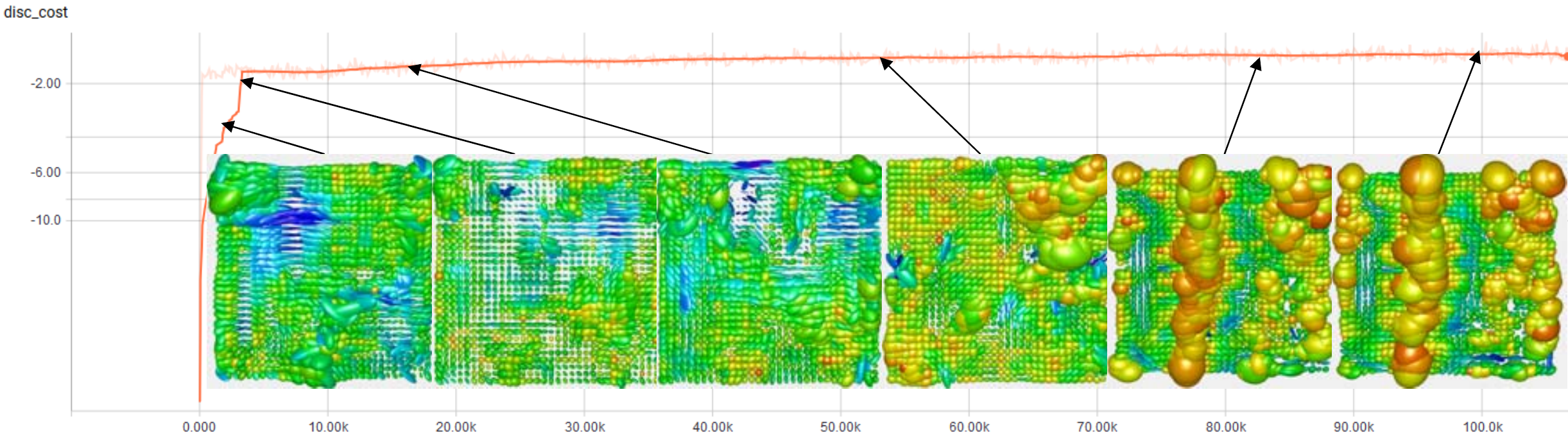}
		\end{center}
		\caption{Training curves and generated samples at different stages of training of the proposed manfioldWGAN on UCL DT image set. The values on y-axis and x-axis denote the negative critic cost (Wasserstein distance) and the iteration number respectively.}
		\label{fig:conv-dt} 
	\end{figure}

Similar to the last evaluation, in Tab.\ref{lab:fid} we also employ the FID metric to compare our manifoldWGAN against the state-of-the-art methods quantitatively for CIFAR-10 and ImageNet. For the FID computation, we translate CB images back to RGB images. By comparing with the models that work for RGB images directly, we can see both the WGAN-CB and our manifoldWGAN-CB performs relatively worse. This is reasonable because the CB-to-RGB translation is non-trivial and lossy. Nevertheless, our proposed manfioldWGAN-CB can still surpass the real competitor WGAN-CB, showing the clear superiority of our proposed model when working for CB images.

We also show the correlation between the proposed manifold-ware Wasserstein GAN loss and the quality of generated CB images in Fig.\ref{fig:conv_hsv_cb}. From the result, we can observe that the quality of generated CB samples tend to be better as the iteration number increases.

\subsection{DT image Generation}
We employ Camino brain DT image set \cite{cook2006camino} to evaluate our proposed manifoldWGAN for the SPD-valued image generation on $\text{SPD}(3)$. The Camino project \cite{cook2006camino} provides a dataset of a Diffusion Tensor Magnetic Resonance Image (DT-MRI) of human heads, which is freely available\footnote{To obtain the data, one can follow the tutorial at \url{http://cmic.cs.ucl.ac.uk/camino//index.php?n=Tutorials.DTI}}.
The UCL DT image database contains 50 brain DT images in total. For data augmentation, we extract 26,080 DT slices of size $32 \times 32$ from the original DT images. Some ground truth DT slice images are listed in Fig.\ref{fig:ucl-dt} (a). Among the sampled DT slice images, there are around 0.6\% non-SPD voxels in each DT slice on average, and only 2,399 samples contain no non-SPD voxels.

For generating visually favorable DT images, our manifoldWGAN suggests to introduce logarithm map and exponential map Eqn.\ref{Eq7} (where each pixel value of the reference point $\bm{y}$ is set to the identity matrix of size $3\times 3$) on SPD manifolds to the Wasserstein GAN setting. In contrast, the original WGAN treats the SPD data as real-valued data. In order to make the generated samples of WGAN to become valid DT (SPD-valued) images, we additionally employ our presented exponential map to transform the outputs of WGAN to the data on SPD manifolds. In Fig.\ref{fig:ucl-dt}, we quantitatively compare these two different approaches on the UCL DT image set. Note that, to plot the generated DT images, we leverage the Manifold-valued Image Restoration Toolbox\footnote{The toolbox is available at \url{http://www.mathematik.uni-kl.de/imagepro/members/bergmann/mvirt/}.}. From the results, we can discover that the generated samples of WGAN is not plausible, while our manifoldWGAN can often synthesize highly realistic DT samples, which would be very useful to the field of DT-MRI that generally lacks of DT images for analysis.

As the last two evaluations, we also study the relationship between the proposed manifold-aware Wasserestein loss and the generated sample quality in Fig.\ref{fig:conv-dt}. As the network training iterations increases, the negative Wasserstein loss goes up quickly and the visual quality of the generated DT images is also improved, again showing there are clear correlations among our manifoldWGAN loss, the convergence property and the quality of its generated samples.

\section{Discussion and Outlook}

We introduced a new generative modeling problem over manifold-valued images, which are often encountered in a wide variety of applications including HSV/CB color and DT-MRI image processing. 
To address the new problem, we generalize the Wasserstein GAN methodology to manifold-valued generation by introducing the manifold geometry-aware optimal transport theory. We finally suggested three benchmarks to evaluate the generative power of our proposed method, and clearly show the superiority of our approach over the state-of-the-art Wasserstein GAN.

For future works, inspired by \cite{shrivastava2017learning} that leverage generative modeling to improve other visual tasks, we will further exploit the practical benefits of our proposed generative models to a broad number of classical manifold-valued data processing tasks like denoising, inpainting and segmentation. In addition, we hope our work could open the path for more manifold-valued image generation tasks like shape and Grassmann manifold-valued data generation. Last but not least, like real-valued image generation, we also encourage more works to extend our framework to conditional settings and even semi-supervised learning.

\noindent\textbf{Acknowledgement:} We would like to thank Nvidia for donating the GPUs used in this work.

%{\small
%	\bibliographystyle{aaai}
%	\bibliography{egbib}
%}

\end{document}